\documentclass[11pt,a4paper]{article}
\usepackage[T1]{fontenc}
\usepackage[utf8]{inputenc}
\usepackage{lmodern}
\usepackage[margin=1in]{geometry}
\usepackage{amsmath,amssymb}
\usepackage{graphicx}
\usepackage{booktabs,array}
\usepackage[font=small,labelfont=bf]{caption}
\usepackage{microtype}
\usepackage[hidelinks]{hyperref}
\usepackage[section]{placeins}
\hypersetup{pdftitle={PERA: Plug-in Experts for Robotic Actions in Continual Whole-Body Control},pdfauthor={Shangchao Hu, Yadan Sun, Jacky Yu, David Huang},pdfsubject={Humanoid robotics and continual whole-body control}}
\title{\bfseries PERA: Plug-in Experts for Robotic Actions\\in Continual Whole-Body Control}
\author{%
  Shangchao Hu\\{\small\href{mailto:shangchao.hu@tcl.com}{shangchao.hu@tcl.com}} \and
  Yadan Sun\\{\small\href{mailto:yadansun@tcl.com}{yadansun@tcl.com}} \and
  Jacky Yu\\{\small\href{mailto:yuxibo@tcl.com}{yuxibo@tcl.com}} \and
  David Huang\\{\small\href{mailto:huang.w@tcl.com}{huang.w@tcl.com}}%
}
\date{}

\begin{document}
\maketitle
\begin{abstract}
Motion imitation enables humanoid robots to acquire whole-body skills such as walking and dancing. However, fine-tuning an entire policy for a new skill can affect existing behaviors, while training a separate policy for each skill increases training and maintenance costs. We present PERA, a method for learning new skills by reusing existing motion capabilities. PERA freezes a pretrained mixture-of-experts policy and trains an additional task expert and its routing score using task rewards or teacher-action supervision. In Unitree G1 simulation, each added task module contains approximately 1.03\% of the parameters of the base actor. Built on our pretrained policy, PERA successfully completes all ten new tasks. Disabling the expert restores the original computation path of the base policy, which still fully executes all 126 original motions. Enabling the expert also improves the execution of some dance motions not used to train that expert, indicating that its benefits extend beyond its training motions. Experiments with HoloMotion further demonstrate the applicability of PERA to different base policies.
\end{abstract}

\section{Introduction}
Reference-based motion imitation provides an effective approach to humanoid whole-body control. With advances in motion data and policy learning, a single controller can perform walking, dancing, and a range of highly dynamic motions~\cite{deepmimic,phc,sonic}. However, new tasks may involve contact patterns, posture constraints, or motion objectives not covered during training. Existing motion-tracking capabilities alone do not guarantee success on these tasks. It therefore remains necessary to learn new skills while preserving access to existing motion capabilities.

\begin{figure}[htbp]
\centering
\includegraphics[width=\linewidth]{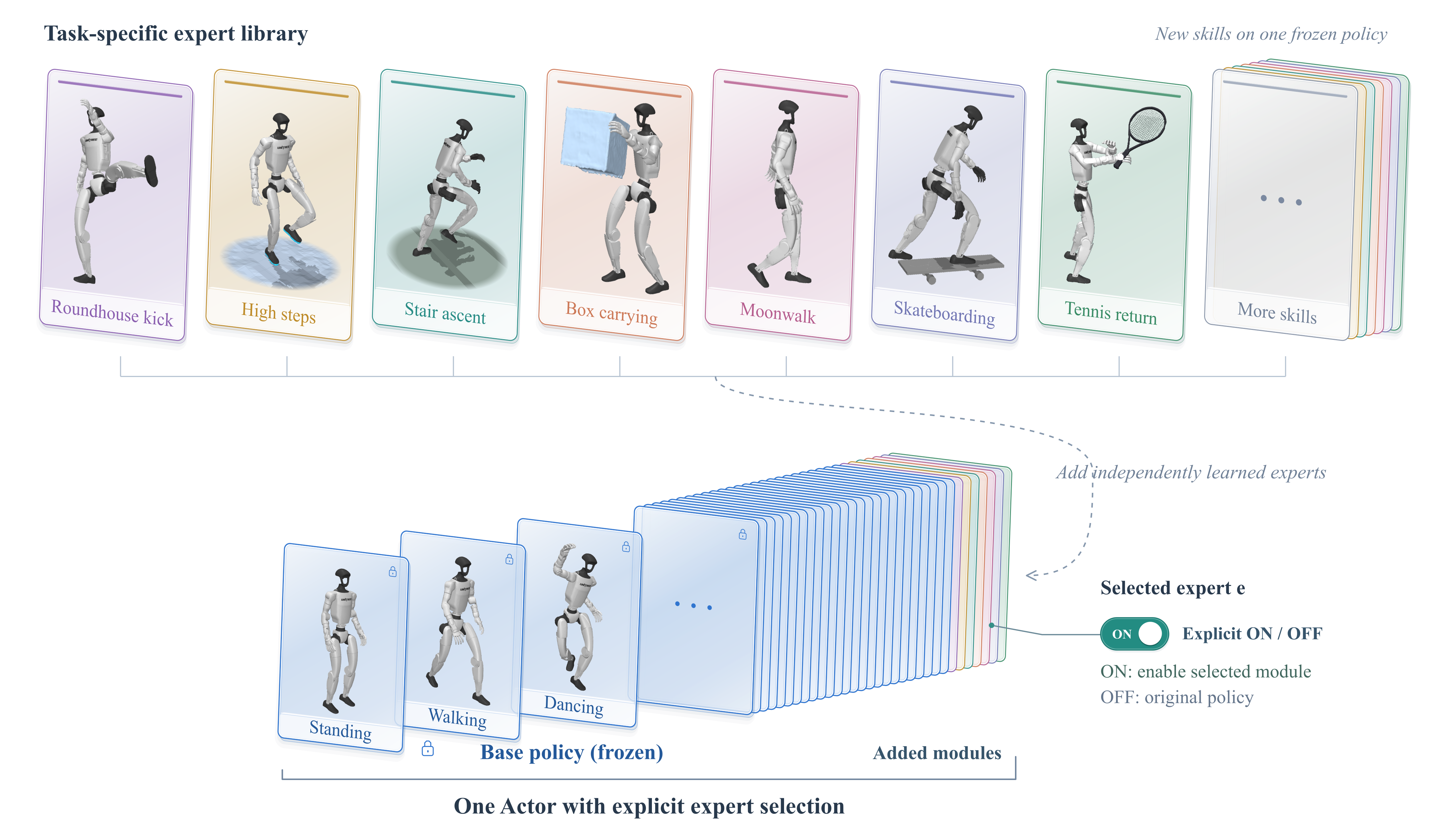}
\caption{\textbf{Skill expansion on a frozen base policy.} Task experts are learned independently and plugged into the same controller on demand, while the base-policy parameters remain frozen. An explicit switch determines whether the selected expert participates in execution. Disabling the expert restores the original computation path.}
\label{fig:overview}
\end{figure}

Directly fine-tuning an entire policy applies new-task gradients to shared parameters and can change existing control behaviors. Maintaining a separate controller for each new task can isolate parameter updates, but requires repeated model storage and makes it difficult to share computation during execution. Skill expansion therefore calls for both reuse of existing motion knowledge and independent access to the original policy. We address this objective by combining a frozen base policy with additional task modules (Figure~\ref{fig:overview}).

Mixture-of-experts (MoE) networks provide a structural basis for incremental expansion. Experts reuse shared features and an action-output interface, while routing determines their participation and mixing weights from the input. Task-related feature transformations can thus be learned within an additional module. A new candidate can also change the selection and mixing of existing experts, so control outputs may change even when shared parameters remain fixed. Expert expansion must therefore specify both how additional modules are learned and how the original policy path is recovered when they are disabled.

PERA learns task experts independently on a frozen MoE base policy, updating only the additional expert and its routing score. Training and execution use the same competitive routing, allowing a task module to learn while participating in control. We use proximal policy optimization (PPO) when task rewards are available, and behavior cloning with online DAgger when reliable teachers are available. During execution, an explicit switch enables the chosen expert on demand; disabling it bypasses the additional module and restores the original computation path.

Our main contributions are: (1) an expert-expansion architecture and training method for continual robot skill learning, in which new experts reuse existing motion representations while the pretrained policy remains frozen; (2) an evaluation of the improvements provided by expert learning, showing higher task completion and improved tracking accuracy on some unseen motions; and (3) a modular skill library that supports independent training, storage, and on-demand invocation of task experts, allowing the skill set to grow through incremental module additions.

\section{Related Work}
\paragraph{Whole-body motion control.}
DeepMimic~\cite{deepmimic} combines reference-motion imitation with task rewards to learn dynamic skills such as walking, rolling, and martial arts through reinforcement learning. PHC~\cite{phc} progressively increases policy capacity to learn large motion collections and incorporates fall recovery, enabling continued execution under noisy references and unexpected loss of balance. SONIC~\cite{sonic} scales motion tracking along model capacity, motion data, and compute, and connects teleoperation and motion-planning interfaces through a unified motion representation, demonstrating the role of large-scale motion learning in general whole-body control. KungfuBot2~\cite{kungfubot2} uses an orthogonal mixture-of-experts architecture to promote distinct motion representations, together with hybrid local--global tracking objectives and a segment-level reward to improve multi-skill and long-horizon motion control. This work studies versatile skills within a unified policy; PERA focuses on adding independently learned task experts after freezing a pretrained policy.

\paragraph{Continual skill learning.}
Similarity-guided LoRA-PNN~\cite{continualhumanoid} selects reusable historical skills based on motion similarity and combines low-rank adaptation with progressive network expansion to learn new motions, retaining historical branches while facilitating transfer across skills. \mbox{LiMoDE}~\cite{limode} addresses lifelong robot manipulation by first learning dynamic experts through multitask pretraining and then combining new and frozen experts for new tasks. DMPEL~\cite{dmpel} progressively constructs a low-rank expert library, dynamically combines experts with a lightweight router, and uses expert coefficient replay to preserve expert invocation on historical tasks, balancing adaptation and knowledge retention.

\paragraph{Expert modules and knowledge integration.}
MoE-Loco~\cite{moeloco} uses expert specialization to alleviate gradient conflicts in multitask reinforcement learning, enabling a single policy to handle multiple terrains and gaits while exhibiting specialized expert behaviors. FlexOlmo~\cite{flexolmo} addresses language-model learning when training data cannot be shared: data owners train experts independently, then combine them through domain-informed routing, allowing modules to be included or excluded at inference time. CPC~\cite{cpc} separates skill acquisition from policy consolidation by training task teachers independently through reinforcement learning, continually distilling their behaviors into a unified student, and mitigating forgetting with an expandable expert architecture and prioritized trajectory replay.

\paragraph{Pretrained control policies and model-assisted learning.}
HoloMotion-1~\cite{holomotion} trains a sparse MoE Transformer on video-reconstructed motions, motion-capture data, and other motion sources. Temporal modeling, sparse expert activation, and sequence-level training improve whole-body tracking, with evaluations of unseen-motion generalization and real-robot transfer. SuperMimic~\cite{supermimic} jointly learns a differentiable world model and a control policy, optimizing the policy by backpropagating through short predicted trajectories to reduce the training cost of large-scale motion tracking.

\section{Method}
\begin{figure}[htbp]
\centering
\includegraphics[width=\linewidth]{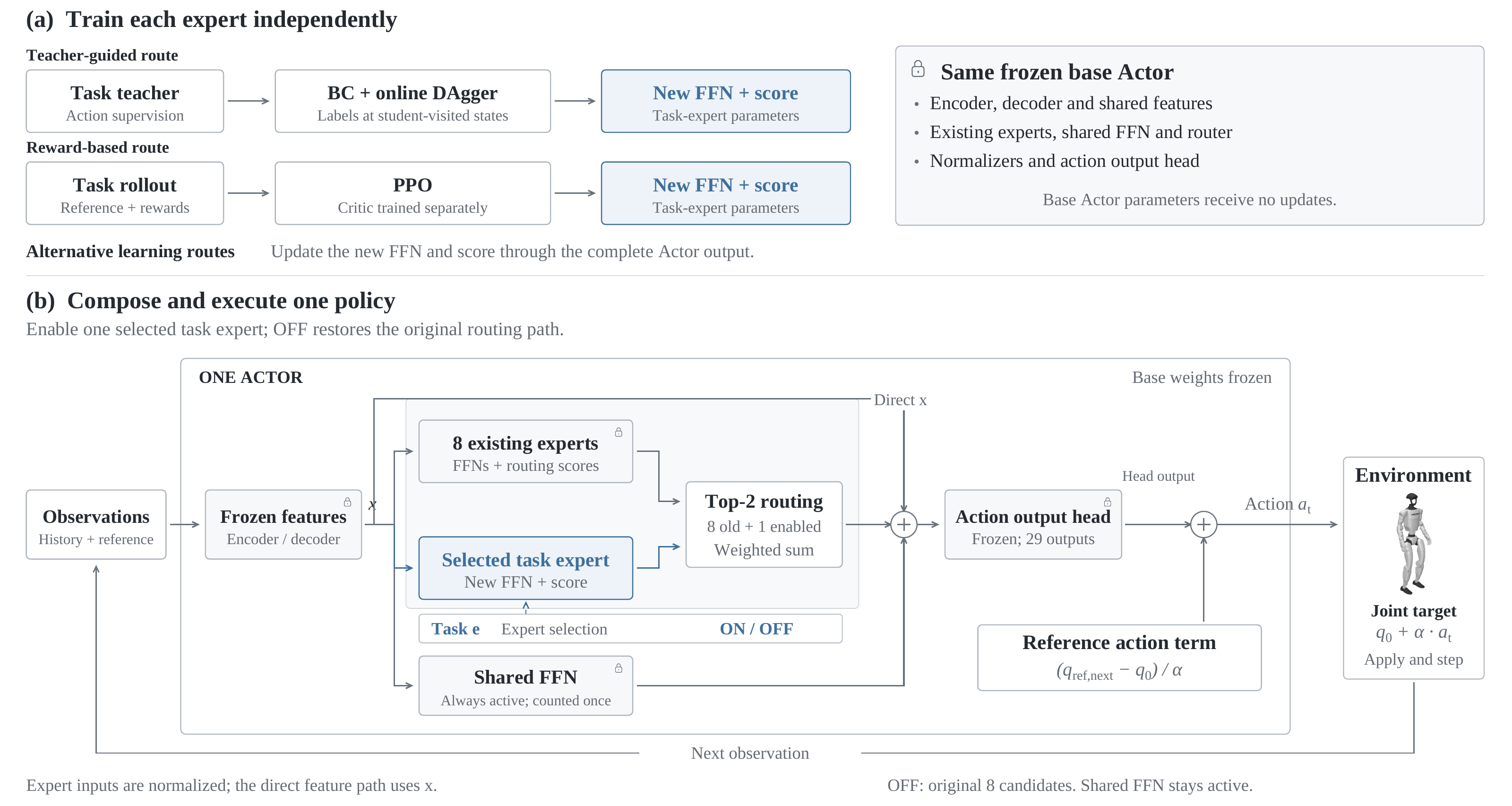}
\caption{\textbf{Learning and executing a task expert inside a frozen actor.} (a) PPO, or behavior cloning followed by online DAgger, updates only the additional FFN and routing score. (b) The eight existing routed experts and one enabled new expert jointly participate in Top-2 selection; the shared FFN is combined separately. Disabling the new expert restores the original routing.}
\label{fig:method}
\end{figure}

PERA first pretrains a base policy on reference motions, then freezes its parameters and statistics and adds a feedforward network (FFN) expert and a scoring branch for each new task. The new and existing experts share the action-output head. Task rewards or teacher supervision update only the additional parameters. Figure~\ref{fig:method} illustrates learning and execution.

\subsection{Pretrained Whole-Body Motion-Tracking Policy}
The base policy follows the reference-motion encoder and observation-history decoder structure of SuperMimic~\cite{supermimic}. The encoder extracts target features from a normalized reference-motion window $g_t$, and the decoder combines them with the most recent ten normalized proprioceptive observations $o_{t-9:t}$ to produce control features:
\begin{equation}
z_t=\operatorname{Enc}(g_t),\qquad
x_t=\operatorname{Dec}_{\mathrm{feat}}(o_{t-9:t},z_t),
\label{eq:encode}
\end{equation}
where $\operatorname{Dec}_{\mathrm{feat}}$ denotes the feature-producing part of the decoder. The complete policy, including expert modules and the action-output mapping, is written compactly as
\begin{equation}
a_t=\pi_\theta(o_{t-9:t},g_t),
\label{eq:action}
\end{equation}
where $a_t\in\mathbb{R}^{29}$ is the control action. Shared and routed experts transform the control features before the unified action-output head. Their selection and mixing are described in Section~\ref{sec:expertlearning}.

Policy pretraining alternates simulator sampling, world-model fitting, and policy updates. A differentiable world model learns dynamics from collected state transitions, and the actor is updated using gradients of predicted returns over 32 steps. After pretraining, the existing experts, router, encoder, output head, and normalization statistics are frozen.

\subsection{Task-Expert Learning}\label{sec:expertlearning}
The FFN of an additional task expert is initialized by copying parameters from an existing expert, and then learns the new task while the base policy remains frozen.

Training updates only the additional FFN and scoring branch; the base-policy parameters and normalization statistics remain frozen. Losses are computed from the actions of the expanded policy, and gradients propagate through the frozen output head to the additional module.

Experts receive the normalized decoder feature $\bar{x}_t$. The router uses the frozen routing-feature embedding $h_t$; task context $c_t$ is also supplied to the new expert and its scoring branch when required.

The router assigns each candidate expert a raw score $\ell_i$. Adding a fixed bias $\delta_i$ gives the selection score $v_i$:
\begin{equation}
v_i=\ell_i+\delta_i,
\label{eq:routing}
\end{equation}
The existing experts retain their pretrained biases, and the new expert's bias is determined before task training. All selection biases remain fixed during task training and testing.

When enabled, the new expert joins the eight existing candidates. The two experts with the highest $v_i$ are selected; a tie does not replace an existing expert with the new one. Let $I$ denote the selected pair. Their mixing weights $\omega_i$ are obtained by applying softmax to their raw scores $\ell_i$, without the selection biases. All unselected routed experts have zero weight.
The weighted sum of the routed-expert outputs is
\begin{equation}
M(\bar{x},h,c)=\sum_{\substack{i\in I\\i\ne n}}\omega_iE_i(\bar{x})
+\omega_nE_n(\bar{x},c),
\label{eq:mixture}
\end{equation}
where the weight of an unselected expert is zero. The shared expert $E_{\mathrm{sh}}$ is always active and is combined separately from the routed-expert mixture inside the complete policy.

\paragraph{Reward-based learning.}
When a task has an explicit reward function, the expanded policy collects state, action, and reward sequences in simulation. PPO~\cite{ppo} updates the additional FFN and scoring branch using these samples. The value network is trained separately, and the base policy remains frozen.

\paragraph{Teacher supervision.}
When a reliable teacher is available, behavior cloning (BC) first trains the additional module to match the teacher's actions. Online DAgger~\cite{dagger} then uses the teacher to label actions at states visited by the policy; the new samples are added to the training set for continued learning. The probability of executing teacher actions is gradually reduced until the expanded policy executes independently.

\subsection{Expert Invocation and Original-Policy Recovery}
Each task expert independently stores its additional FFN, scoring branch, and fixed selection bias. During execution, the corresponding module can be enabled to participate in Top-2 selection. Disabling it bypasses the module and restores the original routing and action computation of the base policy. Because the base-policy parameters and normalization statistics remain frozen, disabling the module produces the same action as the base policy for the same input.

\section{Experiments}
We first evaluate the original motion capabilities after disabling the added expert and transfer to unseen motions when the expert is enabled. We then compare our pretrained policy and HoloMotion-1~\cite{holomotion} after adding PERA experts on ten scenarios, using task teachers as references.

\subsection{Evaluation Setup and Base Policies}
Experiments use a simulated 29-degree-of-freedom Unitree G1. Our pretrained policy implements an MoE network using multilayer perceptrons (MLPs), with eight routed experts and one shared expert. The actor contains approximately 10.4 million parameters.

The ten new tasks use reference motions and paired terrain assets released with Deep Whole-body Parkour~\cite{deepwholebodyparkour}. They comprise dive rolling, kneel climbing, jump sitting, and roll vaulting, forming ten motion--scene combinations across different terrains and wall settings (S01--S10; Table~\ref{tab:scenarios}).

\begin{table}[htbp]
\centering
\caption{Motion--scene combinations used for the ten new tasks.}
\label{tab:scenarios}
\small
\begin{tabular}{@{}lll@{}}
\toprule
ID & Motion & Scene\\
\midrule
S01 & Dive roll & Laboratory, without walls\\
S02 & Dive roll & Ramp, without walls\\
S03 & Dive roll & Laboratory, with walls\\
S04 & Dive roll & Ramp, with walls\\
S05 & Kneel climb & 50 cm step, without walls\\
S06 & Kneel climb & 50 cm step, with walls\\
S07 & Jump sit & Platform, without walls\\
S08 & Jump sit & Platform, with walls\\
S09 & Roll vault & Platform, without walls\\
S10 & Roll vault & Platform, with walls\\
\bottomrule
\end{tabular}
\end{table}

All policies use the same reference motion and terrain for each scenario. The control frequency is 50 Hz, and episodes begin at the first reference frame with the native failure conditions retained. An episode is successful if the reference sequence finishes without a failure termination. Evaluated policies control the robot independently, without teacher-action execution. Endpoint position error is the distance between the root position at episode termination and the final reference root position, including failed episodes. Root tracking error measures the root-position distance to the time-aligned reference during execution. We compare task completion, endpoint position error, and root position error, averaging equally across scenarios (Table~\ref{tab:positions}). The additional expert module for our pretrained policy contains approximately 1.03\% of the base actor's parameters.

Joint-angle tracking error is measured by root mean square error (RMSE):
\begin{equation}
\operatorname{RMSE}_q=\sqrt{\frac{1}{TJ}\sum_{t=1}^{T}\sum_{j=1}^{J}
\left(q_{t,j}-q_{\mathrm{ref},t,j}\right)^2}.
\label{eq:rmse}
\end{equation}
Here $q_{t,j}$ and $q_{\mathrm{ref},t,j}$ are the actual and reference joint angles, respectively, and $J=29$. To use the same comparison interval, we include only the first $T$ frames during which neither policy has terminated. Errors are reported in degrees.

\subsection{Capability Retention and Transfer}
Disabling the additional expert restores the base policy's computation path and produces identical actions for identical inputs. Both the base policy and the controller with the expert disabled fully execute all 126 original motions, with similar mean returns and tracking errors (Table~\ref{tab:retention}).

\begin{table}[htbp]
\centering
\caption{Capability retention and mean tracking errors on original motions.}
\label{tab:retention}
\small
\begin{tabular}{@{}lrr@{}}
\toprule
Metric & Base policy & Expert disabled\\
\midrule
Fully completed motions & 126 & 126\\
Mean normalized return & 0.7893 & 0.7892\\
Mean per-frame joint RMS (${}^\circ$) & 8.0640 & 8.0635\\
Mean per-frame body RMS (cm) & 5.5674 & 5.5554\\
\bottomrule
\end{tabular}
\end{table}

\begin{figure}[htbp]
\centering
\includegraphics[width=\linewidth]{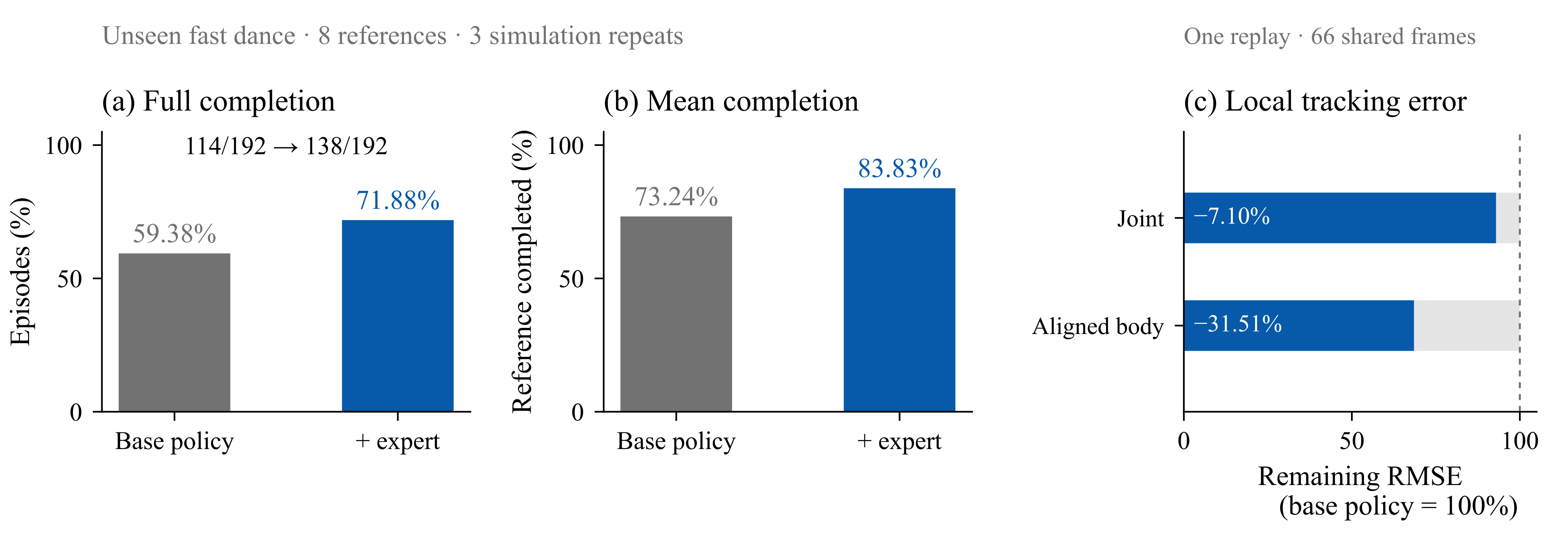}
\caption{\textbf{Execution of dance motions not used to train the expert.} (a) Fraction of episodes that fully execute the reference motion. (b) Mean motion completion. (c) One replay of the motion with the largest completion improvement: joint-angle and aligned-body-position RMSE are compared over the first 66 frames during which neither policy has terminated. Base-policy error is normalized to 100\%; blue bars show the relative error with the expert enabled.}
\label{fig:transfer}
\end{figure}

To test whether the capabilities provided by an expert extend beyond its training motions, we compare execution with the expert disabled and enabled on eight fast-dance references not used to train that expert. No policy parameters are updated during testing. Enabling the expert increases the fraction of episodes that fully execute the reference from 59.38\% to 71.88\%, and mean motion completion from 73.24\% to 83.83\% (Figure~\ref{fig:transfer}(a,b)). These results show that the improvements provided by the expert can extend beyond its training motions.

\subsection{Expert Expansion on Different Base Policies}
To evaluate PERA on different base policies, we add task experts to our pretrained policy and to HoloMotion, then compare endpoint and root position errors on the same ten tasks, with task teachers as references.

\begin{figure}[htbp]
\centering
\includegraphics[width=\linewidth]{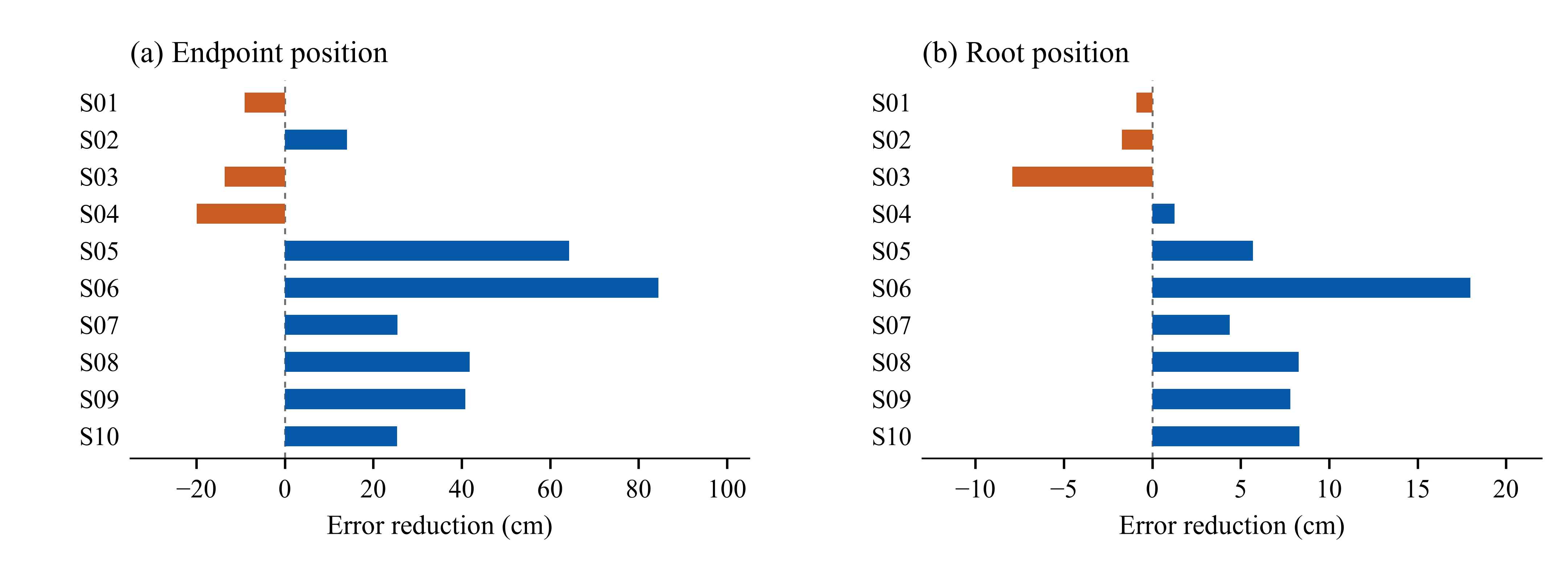}
\caption{\textbf{Per-task errors after expert expansion on different base policies.} (a) Endpoint position error. (b) Root position error. Each bar gives the error of PERA on our pretrained policy minus that of HoloMotion + expert. Positive values indicate lower error for HoloMotion + expert; negative values indicate lower error for PERA on our pretrained policy.}
\label{fig:basecomparison}
\end{figure}

After adding task experts to HoloMotion, mean endpoint position error across the ten tasks decreases from 173.36~cm to 28.51~cm, and mean root position error decreases from 15.12~cm to 14.37~cm. HoloMotion + expert has lower mean error on both metrics than PERA on our pretrained policy (Table~\ref{tab:positions}), although the per-task results vary (Figure~\ref{fig:basecomparison}).

\begin{table}[htbp]
\centering
\caption{Mean position errors on ten tasks.}
\label{tab:positions}
\small
\begin{tabular}{@{}lrr@{}}
\toprule
Policy & Endpoint error (cm) & Root error (cm)\\
\midrule
HoloMotion only & 173.36 & 15.12\\
PERA (our pretrained policy) & 53.88 & 18.69\\
HoloMotion + expert & 28.51 & 14.37\\
Task teacher & 27.79 & 14.87\\
\bottomrule
\end{tabular}
\end{table}

\begin{figure}[htbp]
\centering
\includegraphics[width=\linewidth]{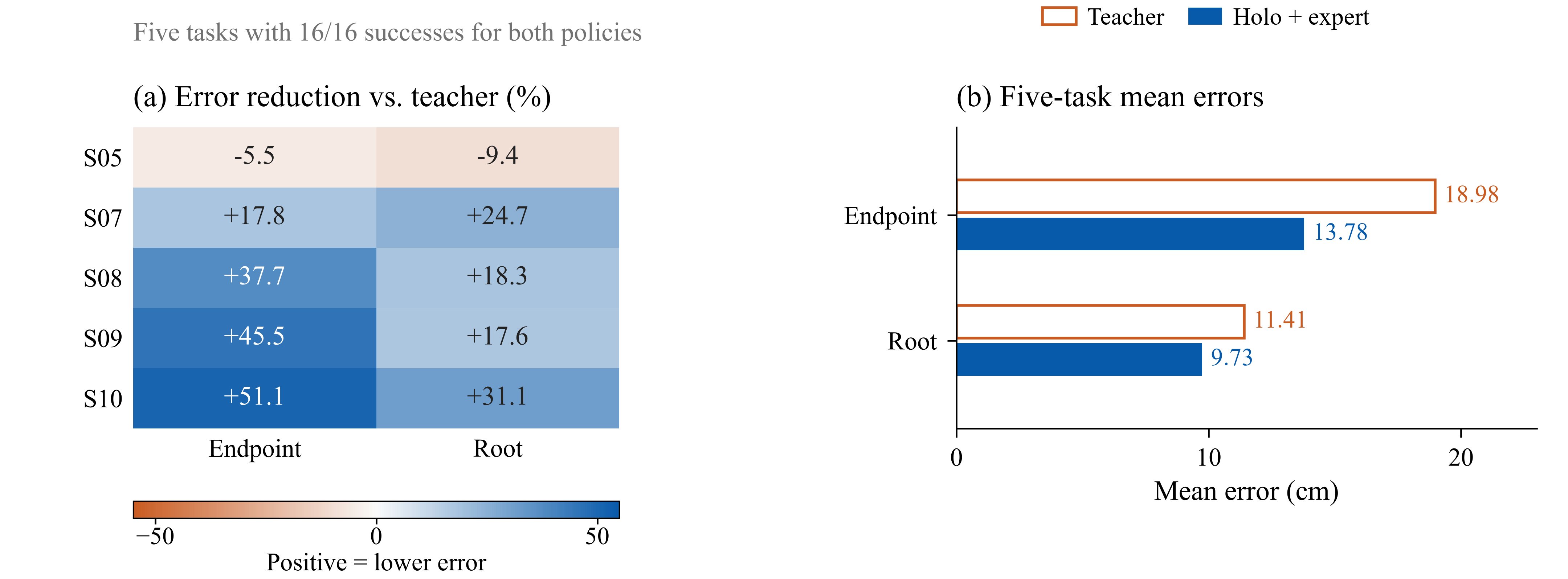}
\caption{\textbf{Position errors of HoloMotion + expert and task teachers.} The comparison includes the five tasks on which both policies fully complete all 16 evaluation episodes. (a) Per-task position-error reductions relative to the teacher: positive values indicate lower error and negative values indicate higher error than the teacher. (b) Position errors averaged equally across the five tasks.}
\label{fig:teachercomparison}
\end{figure}

We further compare the task teacher and HoloMotion + expert on the five tasks for which both policies fully complete all 16 evaluation episodes. The mean endpoint and root position errors of HoloMotion + expert are 27.40\% and 14.70\% lower than those of the teacher, respectively. Both errors are lower than the teacher's on four tasks and higher on the remaining task (Figure~\ref{fig:teachercomparison}).

\section{Conclusion}
We introduced PERA, which expands robot skills by adding task experts to a frozen pretrained motion policy. Each additional expert and its scoring branch are trained independently and participate in action generation alongside existing experts. Experts can be enabled on demand during execution; disabling them restores the original computation path of the base policy, supporting both new-task learning and access to the original policy.

In Unitree G1 simulation, PERA built on our pretrained policy successfully completes all ten new tasks, while disabling the expert preserves full execution of the 126 original motions. Enabling an expert also improves some dance motions not used to train it, showing that its effects can extend beyond its training motions. Experiments with HoloMotion further demonstrate applicability to different base policies. These results support reusing existing motion capabilities and expanding robot skills through independently trained task experts that can be plugged in on demand.

\clearpage

\end{document}